\documentclass{article}

\usepackage{PRIMEarxiv}
\usepackage[utf8]{inputenc} 
\usepackage[T1]{fontenc}    
\usepackage{hyperref}       
\usepackage{url}            
\usepackage{booktabs}       
\usepackage{amsfonts}       
\usepackage{nicefrac}       
\usepackage{microtype}      
\usepackage{lipsum}
\usepackage{fancyhdr}       
\usepackage{graphicx}       
\usepackage{amsmath}
\usepackage{times}
\usepackage{multirow} 
\graphicspath{{media/}}     

\title{L-VITeX: Light-weight Visual Intuition for Terrain Exploration  
}

\author{
  Antar Mazumder \\
  Dept. of Mechatronics Engineering \\
  Rajshahi University of Engineering \& Technology (RUET) \\
  Rajshahi 6204, Bangladesh\\
  \texttt{antar.mte@ieee.org} \\
   \And
  Zarin Anjum Madhiha \\
  Dept. of Electrical and Electronic Engineering \\
  Brac University \\
  Dhaka 1212, Bangladesh\\
  \texttt{zarin.anjum.madhiha@g.bracu.ac.bd} \\
}

\begin{document}
\maketitle

\begin{abstract}

This paper presents L-VITeX, a lightweight visual intuition system for terrain exploration designed for resource-constrained robots and swarms. L-VITeX aims to provide a hint of Regions of Interest (RoIs) without computationally expensive processing. By utilizing the Faster Objects, More Objects (FOMO) tinyML architecture, the system achieves high accuracy (>99\%) in RoI detection while operating on minimal hardware resources (Peak RAM usage < 50 KB) with near real-time inference (<200 ms). The paper evaluates L-VITeX's performance across various terrains, including mountainous areas, underwater shipwreck debris regions, and Martian rocky surfaces. Additionally, it demonstrates the system's application in 3D mapping using a small mobile robot run by ESP32-Cam and Gaussian Splats (GS), showcasing its potential to enhance exploration efficiency and decision-making.

\end{abstract}


\section{Introduction}
Before we delve into the technical jargon, let us first set our problem premises so that we can understand the solution better. There is an interesting similarity between a human going through a long text and a robot exploring a terrain collecting visual data - much of the information is just background noise i.e., they are not actually what the human or the robot is looking for. For example, one might want to learn about Gaussian Splats(GS)\cite{kerbl20233d} in an entire book on 3D computer vision. But, a lot of the chapters or elements in a chapter in the book could be not relevant to GS at all. In such cases, what do we humans do? --- We could either 1) \textit{search for keywords or topic headings on GS and completely skip the others} or 2) \textit{skim through the entire book to get an overall idea but mostly focus on the sections with keywords and headings related to GS}. In the case of a robot explorer, we can consider the instance of an Autonomous Underwater Vehicle (AUV) exploring a region with a mission to monitor coral reefs. Since monitoring corals is its primary objective, it should spend more time inspecting them instead of allowing equal focus to the entire scene- and that is where the requirement for Region of Interest (RoI) detection comes into play. In humans, such emphasis is a cognitive focus, in machines, however, focus is preceded by detection\cite{liu2020lightweight} which is presently performed through machine learning or deep learning algorithms on 2D images or depth feeds. 

The most cutting-edge visual guidance methods such as the ones utilized in the DARPA Subterranean Challenge typically leverage depth data from time-of-flight sensors such as LiDARs, which not only enable better object distance estimation but also object shape estimation-facilitating classification\cite{miles2023terrain}. This type of data is also viable in classifying access zones in multi-robot terrain exploration as demonstrated by Ulloa \textit{et al} in\cite{christyan_cruz_ulloa__2024} achieving 95\% success rate. Extending scope, depth vision has also been leveraged in egocentric vision-based guidance for terrain estimation in legged robots\cite{agarwal2023legged}. Furthermore, in many cases, depth data from a LiDAR sensor has been merged with 2D image input through sensor fusion\cite{shaukat2016towards}, such as the VA-LOAM architecture in\cite{taeyoung_jung__2024}. In others, the two inputs have been combined to construct surface normal images to be processed by deep learning models to estimate navigable space\cite{chen2021navigable}. On the contrary, monocular vision-based methods in facilitating robot guidance are usually of two types- one-stage (YOLO variants, SSD, etc.) and two-stage(CNN, R-CNN, etc.), the former being the most popular due to fast real-time performance\cite{tang2023survey}. The most widespread application in such cases is mostly object detection and semantic segmentation. For example, Guo \textit{et al.} performed Martian terrain classification using YOLOv5 in\cite{guo2023terrain} for planetary explorers. In another research, Lee \textit{et al.} devised semantic segmentation for robot path planning from visual data using PSPNet50 based on the traditional ResNet50 architecture\cite{lee2024terrain}. Beyond object detection, recognition, and segmentation, vision input has also been mapped to detect physical parameters of terrain for exploration and navigation as demonstrated in\cite{chen2024identifying} where terrain images have been used to determine surface friction with Mean Absolute Error (MAE) of 0.15. In modern times, considering the ubiquity of monocular cameras, methods such as Structure from Motion (SfM)  and Gaussian Splatting\cite{matsuki2024gaussian} have been on a steady rise compensating for the lack of depth information in 2D image data. GS, which is a post-process of SfM can provide a 3D reconstructed map for facilitating robot mapping and navigation. For example, in\cite{tao2024rt} Tao \textit{et al.} demonstrated real-time navigation from an information-rich GS. Similar GS-based maps have also been leveraged in\cite{chen2024splat} for robot navigation. 

However, while such algorithms are suitable to deploy into large primary explorers like large tethered AUVs or UGVs, they are either incompatible with resource-constrained hardware or perform very poorly in the same. Moreover, while large explorers, with their onboard computation power, can perform very complex tasks such as real-time 3D mapping of exploring terrains, they all suffer from poor maneuverability in restricted spaces such as underground caves or exo-planet lava tubes. In such settings, small-secondary explorers could be viable options but due to their small footprint, they cannot carry bulky and powerful computation hardware; thus, limiting this scope of application. In addition, even if the explorer is large enough to carry some powerful computing hardware, minimizing weight remains a top priority. This is especially crucial for UAV explorers in less dense environments, such as exoplanets, where more weight directly translates to higher energy requirements for operation. Even Ingenuity, NASA's secondary UAV in the Mars 2020 mission, faced significant design challenges to generate enough lift in the Martian environment, which has only 2\% of the atmospheric density of Earth's~\cite{balaram2021ingenuity}. Therefore, deploying efficient, lightweight hardware can also help overcome such challenges by making the robot lighter. 

\begin{figure}
    \centering
    \includegraphics[width=1\linewidth]{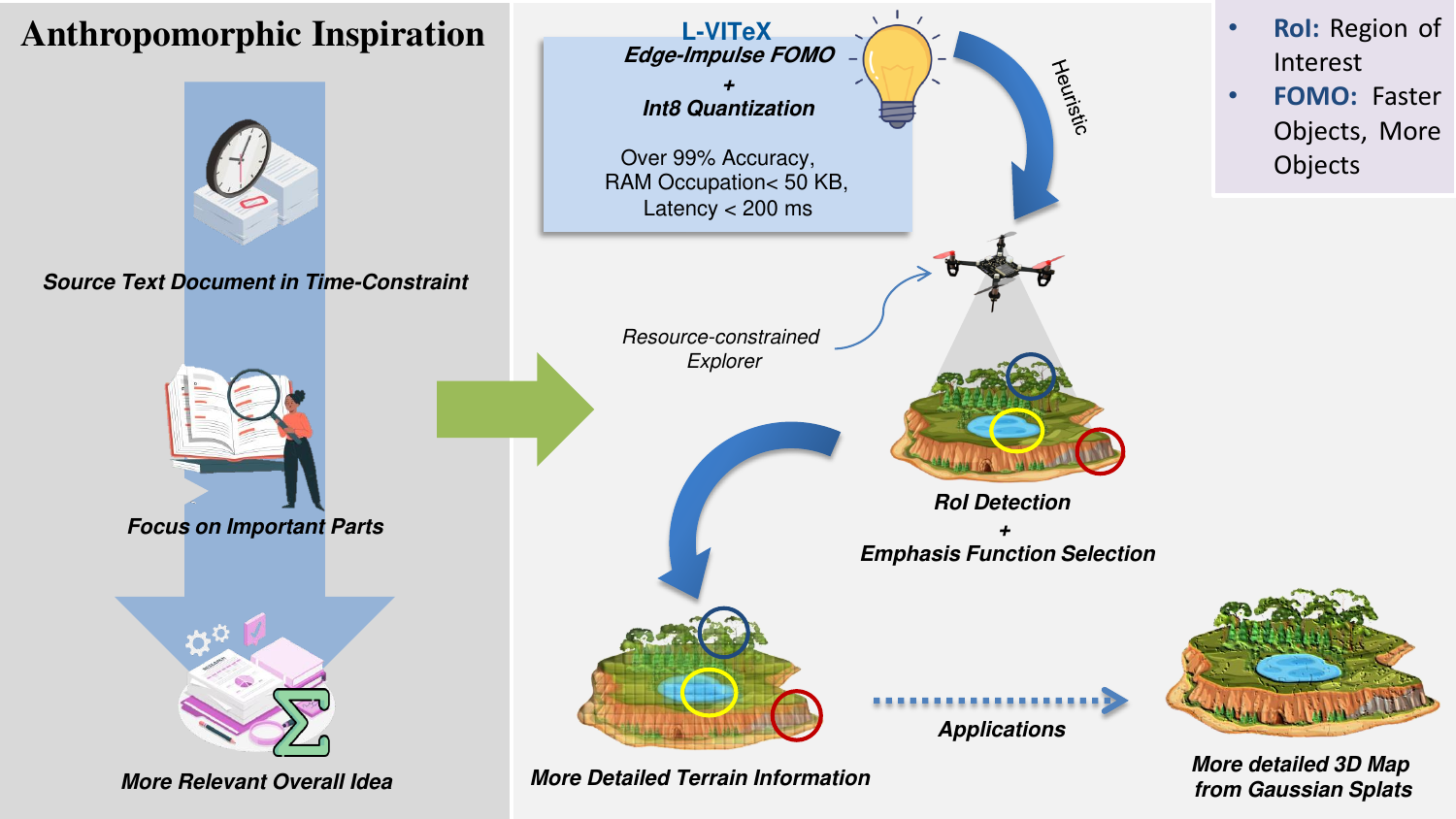}
    \caption{Overview of the visual heuristic process building on anthropomorphic focus.}
    \label{fig: overall}
    
\end{figure}

In recent times, with the emergence of AI on the edge with object detection models such as MobileNet SSD\cite{dong2020mobilenetv2}, the potential to deploy RoI detection in small explorers could be feasible. MobileNetV2\cite{indraswari2022melanoma} is an efficient convolutional neural network optimized for mobile and embedded devices, balancing speed and accuracy\cite{nguyen2020fast}. The introduction of Faster Objects, More Objects (FOMO)\cite{dharani2023object}, a revolutionary tinyML object detection architecture developed by Edge Impulse based on MobileNet V2~\cite{sandler2018mobilenetv2} advances the prospect of running object detection on general-purpose microcontrollers such as ESP-32 with very minimal on-board computational resources\cite{gotthard2023edge}. Although FOMO has not been used for ROI detection for visual navigation of robots, it showed great performance in object detection in resource-limited platforms. For example, Mahur \textit{et al.} showed it to be 30 times faster than MobileNet SSD or YoloV5\cite{mahaur2023small}, making it significantly more efficient for detection tasks\cite{dharani2023object}. This model required considerably less memory, with a footprint that scales to approximately 100KB in RAM.\cite{jiang2021semantic, alaa2024object}. It achieved approximately 3x higher mean Average Precision (mAP) compared to baseline models\cite{zohar2023open}. In\cite{lin2024tiny}, the authors presented a real-time bolt-defect detection system that combines FOMO-TinyML\cite{moosmann2023tinyissimoyolo} for a magnetic climbing robot. The approach detected accurately with a precision of 82\% and an F1 score of 75\%. In \cite{dimitrov2024queen}, the study examined the use of FOMO on Raspberry Pi 5 and Coral Dev Board Micro to automatically detect queen bees. The average accuracy for queen bee detection using the FOMO model on the Raspberry Pi 5 was 86.5\%, while the accuracy of the MobileNetV2 model on the Coral Dev Board Micro was lesser -- 79.91\%. Imron \textit{et al.}\cite{imron2024cloud} explored the use of cloud storage for object detection with ESP32-CAM, optimizing the training of the FOMO model for microcontroller identification, achieving a 98.7\% F1 score and 89.58\% precision. In\cite{boyle2024dsort} the authors proposed adaptive tiling for the detection of lightweight objects in low-power MCUs, improving F1 scores by up to 225\% while maintaining energy efficiency and precision using FOMO networks. Thus, in object detection on the edge, FOMO has shown great performance. 

Addressing the research problems previously discussed and the potential of FOMO to solve them, this study proposes L-VITeX, a lightweight visual Intuition system for Terrain Exploration for small secondary resource-constrained explorers and robot swarms. The goal of the proposed system is to provide a hint of the presence of RoIs without making the process computationally expensive. As a result, the explorer can then select an emphasis function(EF) triggered by the on-board FOMO model, such as traveling slower if RoI is hinted, going down for a closer look in the case of small UAVs and AUVs, changing formation in robot swarms, or just simply avoiding taking videos of irrelevant regions to save memory and so on. Additionally, it addresses the following research questions-

\begin{itemize}
    \item \textbf{RQ1:} \textit{Can the FOMO-empowered L-VITeX be leveraged for near real-time (<1sec) RoI detection for visual heuristic-based terrain exploration by resource-constrained explorers?}
    \item \textbf{RQ2: }\textit{What are the best and the worst case scenarios for applying the approach?}
\end{itemize}

 To test its performance, we validate performance metrics such as F1-score, precision, recall, accuracy, peak ram usage, and inference latency for four different image datasets from three distinct terrain exploration scenarios: 1) aerial mountainous terrain exploration with multiple Regions of Interest (ROIs), 2) underwater exploration for shipwreck debris regions, and 3) Martian surface exploration through rock detection. Furthermore, as a proof of concept, we also showcase the algorithm's application in 3D mapping through GS and compare the maps with and without our proposed visual intuition. In this case, the algorithm operates on a small Espressif ESP32 cam system with a minimal RAM usage below 50 KB while achieving over 99\% accuracy in RoI detection.

The organization of this paper is as follows:  Section 2 describes the dataset, settings, and methodology involved followed by the results and discussion in Section 3. Finally, the paper concludes in Section 4, and future directions are stated.

\begin{table}
    \centering
    \small{
    \begin{tabular}{lcccc} 
        \toprule
        \multicolumn{5}{c}{\textbf{Model Training Parameters}} \\
        \midrule
        Parameters & Drone Videos & Marine Video Kit + Shipwreck & Rocks Detection & Rover and Rocks Detection \\ 
        \midrule
        \multirow{3}{*}{Image Input Size} & 32x32 & 32x32 & 32x32 & 32x32 \\ 
                                          & 64x64 & 64x64 & 64x64 & 64x64 \\ 
                                          & 96x96 & 96x96 & 96x96 & 96x96 \\ 
                                          \midrule
        Train:Test      & 98:24      & 88:23      & 771:194    & 117:33 \\ 
        Learning Rate   & 0.0001     & 0.0005     & 0.0001     & 0.0001 \\ 
        Optimizer       & adam       & adam       & adam       & adam \\ 
        Batch Size      & 32         & 16         & 16         & 32 \\ 
        Number of Epochs& 300        & 300        & 300        & 500 \\ 
        \bottomrule
    \end{tabular}
    }
    \caption{Experimental settings and model training parameters.}
    \label{tab:exp_ds}
\end{table}

\begin{figure}
    \centering
    \includegraphics[width=0.75\linewidth]{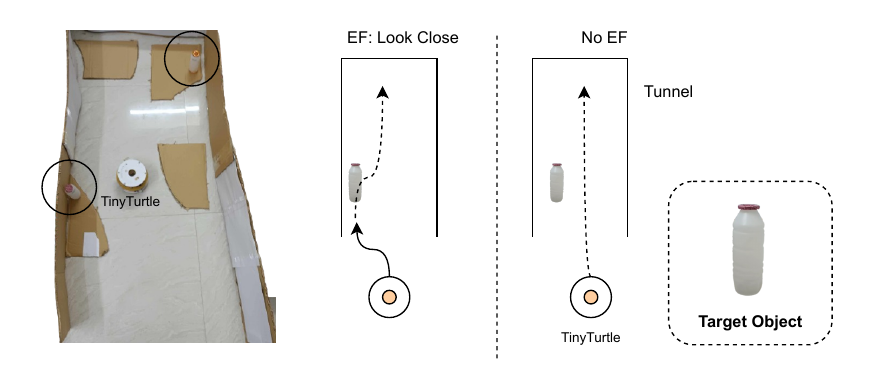}
    \caption{Proof of concept system design and Emphasis Function action.}
    \label{fig:poc}
\end{figure}

\begin{table}
    \centering
    \small{
    \begin{tabular}{lc} 
        \toprule
        \multicolumn{2}{c}{\textbf{Radiance Field Training Parameters}} \\
        \midrule
        Max Image Count & 100 \\
        Max Features/Frame & 8 \\
        Downsample Images & 1024 \\
        Max Splat Count & 3000 \\
        Stop Training after & 30\\
        \bottomrule
    \end{tabular}
    }
    \caption{Radiance field training parameters.}
    \label{tab:exp_gs}
\end{table}

\section{Methodology}
\subsection{Datasets \& Experimental Settings}
There were five datasets involved in training and evaluating the FOMO model variants - \textit{Drone Videos}\cite{mader_drone_videos}(train \& test), \textit{Marine Video Kit}\cite{truong2023marine}(train), \textit{Shipwreck Dataset}~\cite{pecheux2023self}(test), \textit{Rocks Detection}~\cite{rocks-detection-govch_dataset}(train \& test), \textit{Rover and Rocks Detection}~\cite{rover-and-rocks-segmentation_dataset}(train \& test). The first three are video datasets to represent the actual motion of perspectives while exploring relevant terrains. The first represents aerial terrain exploration with a focus on multiple RoIs including ice, vegetation, rocks, and streaks. The next two, especially \textit{Shipwreck} consist of video footage of shipwreck debris on the ocean floor to mimic AUV perspectives. The last two represent exo-planet terrain exploration through rock detection. The \textit{Rocks Detection} dataset comprises actual rock pictures on the Martian surface along with rock images in analogous environments. The \textit{Rover and Rocks Detection} dataset comprises images of an earthly field trial of small explorers and rocks in the field of view. The experimental settings for training the datasets are described in table \ref{tab:exp_ds}.

\subsection{System Architecture}
\subsubsection{The FOMO Model}
The FOMO model utilizes a truncated version of MobileNet-V2\cite{sandler2018mobilenetv2} as its feature extractor\cite{novak2024intelligent}. First, each input image is initially resized to fit the model's requirements. Next, the input image is divided into a grid of fixed dimensions. In our case, we used the default 8x  8 pixels per grid. Thus, a 32x32 image outputs 4x 4 cells, while a 64x64 image produces 8x 8 cells and a 96x96 image can be configured as 12x 12. Each grid cell functions as an independent classification task, and the tasks are carried out in parallel substantially reducing time consumption. Instead of predicting bounding boxes, FOMO focuses on detecting the centroids of objects within each grid cell. The model outputs probability scores indicating the likelihood of an object being present at the centroid of each cell, along with corresponding class predictions. This centroid-based detection simplifies computations and enhances the model's speed. Finally, the model generates a heat map as the output, where each cell corresponds to a probability score indicating the presence of an object and its associated class\cite{fomo}. To make the model resource-efficient we further quantize it by transforming the float32 model weights to 8-bit integers. This sometimes reduces accuracy but drastically reduces resource consumption. 
\subsubsection{Proof of Concept and Emphasis Function}
As a proof of concept, we use TinyTurtle, a low-cost mobile robot running on a single ESP-32 Cam as the robot brain to 3D map a tunnel structure via the Gaussian Splatting process as depicted in figure \ref{fig:poc}. The robot explores through the makeshift tunnel with objects of interest placed at random places. A dedicated FOMO model runs onboard the ESP-32-Cam to provide a visual heuristic while exploring to trigger an emphasis function.
The emphasis function (EF) is the designated robot action to be performed if the on-board FOMO model detects RoI. In our proof-of-concept, we investigated a simple EF for the ESP32-Cam driven mobile robot --- \textbf{Look Close:} \textit{Approach detected RoI for closer inspection.} We run two experiments - GS with and without the proposed FOMO-based L-VITeX and compare the comprehensiveness of the two 3D reconstructions for assessment. The training parameters used for the GS radiance field is described in table \ref{tab:exp_gs}.

\subsection{Performance Metrics}

The performance of the model is evaluated using macro-average precision, recall, F1 score, and accuracy. These metrics provide insights into the model's performance across all classes by averaging the metrics for each class.

\subsubsection{Macro-Average Precision}
Macro-average precision\cite{berger2020threshold} is the average precision across all classes, calculated as:

\begin{equation}
    \text{Macro Precision} = \frac{1}{N} \sum_{i=1}^{N} \frac{TP_i}{TP_i + FP_i}
\end{equation}

where $TP_i$ and $FP_i$ represent the true positives and false positives for class $i$, and $N$ is the total number of classes.

\subsubsection{Macro-Average Recall}
Macro-average recall\cite{berger2020threshold} is the average recall across all classes and is defined as:

\begin{equation}
    \text{Macro Recall} = \frac{1}{N} \sum_{i=1}^{N} \frac{TP_i}{TP_i + FN_i}
\end{equation}

where $TP_i$ and $FN_i$ represent the true positives and false negatives for class $i$.

\subsubsection{Macro-Average F1-Score}
The macro-average F1-score\cite{takahashi2022confidence} is the harmonic mean of macro-average precision and recall, calculated as:

\begin{equation}
    \text{Macro F1} = \frac{1}{N} \sum_{i=1}^{N} 2 \cdot \frac{\text{Precision}_i \cdot \text{Recall}_i}{\text{Precision}_i + \text{Recall}_i}
\end{equation}

\subsubsection{Accuracy}
In Edge Impulse Studio, the accuracy is calculated only for the instances attaining precision over 0.80. However, we considered the generic Accuracy\cite{yacouby2020probabilistic} as the ratio of correctly predicted observations to the total observations, providing an overall effectiveness of the model. It is calculated as:

\begin{equation}
    \text{Accuracy} = \frac{TP + TN}{TP + TN + FP + FN}
\end{equation}

where $TP$ = True Positives, $TN$ = True Negatives, $FP$ = False Positives, and $FN$ = False Negatives.

\begin{table}
    \centering
    \small{
    \begin{tabular}{ccccccccc}
        \toprule
        Dataset & Model Format & Image Size & F1-Score & Recall & Precision & Accuracy & PRO & Latency \\ 
        \midrule
        \multicolumn{9}{c}{\textbf{Aerial Terrain Exploration: Terrain Detection}} \\
        \midrule
        Drone Videos, Kaggle\cite{mader_drone_videos} & f32  & 32x32 & 0.79 & 0.77 & 0.93 & 84.68\% & 50.7 KB & 318 ms\\
                             & f32  & 64x64  & 0.92  & 0.92  & 0.92  & 98.51\%  & 171.5 KB & 1065 ms\\ 
                             & f32  & 96x96  & 0.92  & 0.93  & 0.92  & 99.39\%  & 363.2 KB & 2068 ms\\ 
        \cmidrule(lr){2-9}
                             & int8 & 32x32 & 0.80 & 0.75 & 0.89 & 84.11\% & 41.3 KB & 149 ms\\
                             & int8 & 64x64  & 0.92  & 0.91  & 0.93  & 98.63\%  & 71.5 KB & 531 ms\\ 
                             & int8 & 96x96  & 0.92  & 0.93  & 0.90  & 99.45\%  & 119.4 KB & 1216 ms  \\ 
        \midrule
        \multicolumn{9}{c}{\textbf{Underwater Terrain Exploration: Shipwreck Detection}} \\
        \midrule
        Marine Video Kit\cite{truong2023marine} (Train) & f32 & 32x32 & 0.97 & 0.95 & 0.99 & 99.72\% & 58.7 KB & 509 ms\\
                               & f32    & 64x64 & 0.87 & 0.91 & 0.85 & 99.57\% & 203.5 KB & 1134 ms\\
                               & f32   & 96x96 & 0.93 & 0.95 & 0.91 & 99.91\% & 435.2 KB & 2574 ms\\
        \cmidrule(lr){2-9}
        Shipwreck Dataset\cite{pecheux2023self} (Test) & int8 & 32x32 & 0.97 & 0.95 & 0.99 & 99.72\% & 43.6 KB & 173 ms\\
                               & int8    & 64x64 & 0.93 & 0.91 & 0.95 & 99.79\% & 79.8 KB & 511 ms\\
                               & int8   & 96x96 & 0.99 & 0.95 & 0.87 & 99.80\% & 137.7 KB & 1139 ms\\
        \midrule
        \multicolumn{9}{c}{\textbf{Exoplanet Terrain Exploration: Rock Detection}} \\
        \midrule
        Rocks Detection~\cite{rocks-detection-govch_dataset} & f32  & 32x32 & 0.70 & 0.67 & 0.75 & 83.80\% & 53.2 KB & 673 ms\\ 
                             & f32  & 64x64  & 0.84 & 0.84 & 0.85 & 97.05\% & 225.7 KB & 1224 ms\\ 
                             & f32  & 96x96  & 0.88 & 0.95 & 0.86 & 98.80\% & 443.6 KB & 2549 ms\\ 
        \cmidrule(lr){2-9}
                             & int8 & 32x32 & 0.69 & 0.66 & 0.74 & 83.25\% & 47.7 KB & 186 ms\\ 
                             & int8 & 64x64  & 0.85 & 0.85 & 0.85 & 99.17\% & 83.3 KB & 587 ms\\ 
                             & int8 & 96x96  & 0.87 & 0.89 & 0.85 & 98.76\% & 143.5 KB & 1227 ms\\ 
        \midrule
        Rover and Rocks Detection~\cite{rover-and-rocks-segmentation_dataset} & f32  & 32x32 & 0.63 & 0.59 & 0.82 & 74.81\% & 58.7 KB & 328 ms\\ 
                             & f32  & 64x64  & 0.82 & 0.76 & 0.92 & 95.31\% & 203.5 KB & 1247 ms\\ 
                             & f32  & 96x96  & 0.88 & 0.81 & 0.98 & 98.59\% & 435.2 KB & 1508 ms\\ 
        \cmidrule(lr){2-9}
                             & int8 & 32x32 & 0.64 & 0.59 & 0.82 & 74.62\% & 43.6 KB & 149 ms\\ 
                             & int8 & 64x64  & 0.83 & 0.76 & 0.94 & 95.36\% & 79.8 KB & 513 ms\\ 
                             & int8 & 96x96  & 0.88 & 0.81 & 0.97 & 98.57\% & 137.7 KB & 1152 ms\\ 
        \bottomrule
    \end{tabular}
    }
    \caption{Test performance metrics over different datasets. Here f32 and int8 are the TF Lite 32-bit float and quantized 8-bit integer models respectively. PRO denotes Peak RAM Occupation and Latency is the inference latency on Esp32-Cam.}
    \label{tab:test}
\end{table}

\section{Results \& Discussion}

\begin{figure}
    \centering
    \includegraphics[width=1\linewidth]{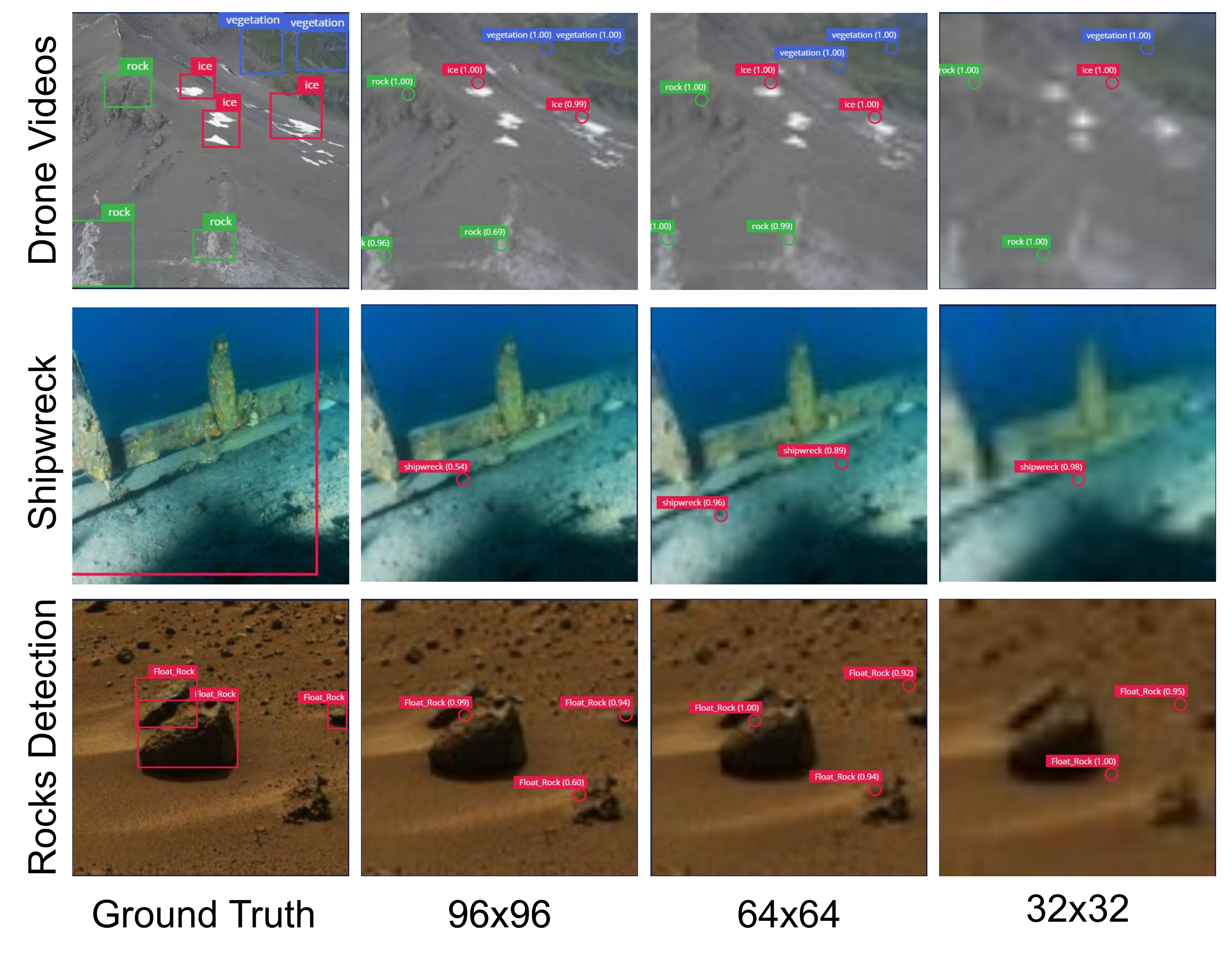}
    \caption{FOMO detection performance across various datasets with three input sizes.}
    \label{fig:datasets}
\end{figure}

\subsection{Object Detection Results}
Table~\ref{tab:test} presents the detailed performance metrics of models evaluated on four different datasets. These models are compared across two formats: 32-bit floating point (f32) and 8-bit integer quantized (int8), and three image resolutions: 32x32, 64x64, and 96x96. The metrics include F1-score, Recall, Precision, Accuracy, Peak RAM Occupation (PRO), and inference latency on the Esp32-Cam, highlighting the tradeoff between model performance and resource consumption.

For the \textit{Drone Videos, Kaggle} dataset, the f32 model at 32x32 resolution achieves an F1-score of 0.79, a Recall of 0.77, and a Precision of 0.93, with an accuracy of 84.68\%. The model has a Peak RAM Occupation (PRO) of 50.7 KB and a latency of 318 ms. Increasing the resolution to 64x64 results in a higher F1-score of 0.92, Recall and Precision both at 0.92, and accuracy of 98.51\%, with a corresponding latency of 1065 ms and PRO of 171.5 KB. At 96x96, the model maintains the F1-score of 0.92 and a slightly higher Recall of 0.93, with a latency of 2068 ms and a PRO of 363.2 KB. The int8 model for the same dataset at 32x32 has an F1-score of 0.80, Recall of 0.75, Precision of 0.89, accuracy of 84.11\%, and a significantly reduced latency of 149 ms with a PRO of 41.3 KB. At 64x64, the int8 model achieves an F1-score of 0.92, with a Recall of 0.91 and a Precision of 0.93, accuracy of 98.63\%, latency of 531 ms, and PRO of 71.5 KB. At the highest resolution, 96x96, the int8 model maintains an F1-score of 0.92 and shows a latency of 1216 ms.

In the \textit{Shipwreck Dataset}, the models were trained on a completely different marine video dataset, \textit{MarineVideoKit} to assess robustness. The f32 model achieves exceptional results at 32x32 resolution, with an F1-score of 0.97, Recall of 0.95, Precision of 0.99, and accuracy of 99.72\%. The latency is recorded at 509 ms, and the PRO is 58.7 KB. At 64x64, the F1-score decreases to 0.87, with a Recall of 0.91, Precision of 0.85, and accuracy of 99.57\%, alongside a latency of 1134 ms and a PRO of 203.5 KB. The 96x96 f32 model achieves an F1-score of 0.93, Recall of 0.95, and Precision of 0.91, with accuracy of 99.91\% and a latency of 2574 ms. In contrast, the int8 model at 32x32 resolution shows an F1-score of 0.97, Recall of 0.95, Precision of 0.99, and accuracy of 99.72\%, with a latency of just 173 ms. At 64x64, the F1-score improves to 0.93, with a Recall of 0.91, Precision of 0.95, and accuracy of 99.79\%, latency of 511 ms, and PRO of 79.8 KB. At 96x96, the int8 model achieves an F1-score of 0.99, Recall of 0.95, and Precision of 0.87, with accuracy of 99.80\%, latency of 1139 ms, and PRO of 137.7 KB.

For the \textit{Rocks Detection} dataset, the f32 model at 32x32 resolution reports an F1-score of 0.70, Recall of 0.67, Precision of 0.75, and accuracy of 83.80\%, with a latency of 673 ms and a PRO of 53.2 KB. At 64x64 resolution, the F1-score increases to 0.84, Recall and Precision both at 0.84, and accuracy of 97.05\%, with a latency of 1224 ms and a PRO of 225.7 KB. At 96x96, the model achieves an F1-score of 0.88, Recall of 0.95, Precision of 0.86, and accuracy of 98.80\%, with a latency of 2549 ms. For the int8 models, the 32x32 version achieves an F1-score of 0.69, Recall of 0.66, Precision of 0.74, and accuracy of 83.25\%, with a latency of 186 ms and a PRO of 47.7 KB. The 64x64 int8 model improves to an F1-score of 0.85, with a Recall and Precision of 0.85, accuracy of 99.17\%, latency of 587 ms, and PRO of 83.3 KB. At 96x96, the int8 model achieves an F1-score of 0.87, Recall of 0.89, and Precision of 0.85, with accuracy of 98.76\% and a latency of 1227 ms.

Finally, for the \textit{Rover and Rocks Detection} dataset, the f32 model at 32x32 resolution achieves an F1-score of 0.63, Recall of 0.59, Precision of 0.82, and accuracy of 74.81\%, with a latency of 328 ms and a PRO of 58.7 KB. At 64x64, the F1-score improves to 0.82, Recall of 0.76, Precision of 0.92, and accuracy of 95.31\%, with a latency of 1247 ms and a PRO of 203.5 KB. At 96x96, the model reaches an F1-score of 0.88, Recall of 0.81, Precision of 0.98, and accuracy of 98.59\%, with a latency of 1508 ms. The int8 model at 32x32 resolution exhibits an F1-score of 0.64, Recall of 0.59, Precision of 0.82, and accuracy of 74.62\%, with a latency of 149 ms and a PRO of 43.6 KB. At 64x64, the F1-score increases to 0.83, Recall of 0.76, Precision of 0.94, and accuracy of 95.36\%, with a latency of 513 ms and a PRO of 79.8 KB. Finally, at 96x96 resolution, the int8 model reaches an F1-score of 0.88, Recall of 0.81, Precision of 0.97, and accuracy of 98.57\%, with a latency of 1152 ms.

Again, in figure \ref{fig:datasets}, the detection performance across various image input sizes for three different datasets representing three distinct terrains can be observed. It is evident that with the reduction of image resolution, the number of detected objects also reduces, except for the shipwreck dataset in the underwater scenario. We also see double detection of a single object by the 64x64 input for the same. The relation between resolution and the number of detected objects is more clear in the other two datasets, especially in Drone Videos. In these cases, we also see a shift of the detected centroid from the target of interest. 

\subsection{Proof of Concept Performance}
\begin{figure}[t]
    \centering
    \includegraphics[width=0.75\linewidth]{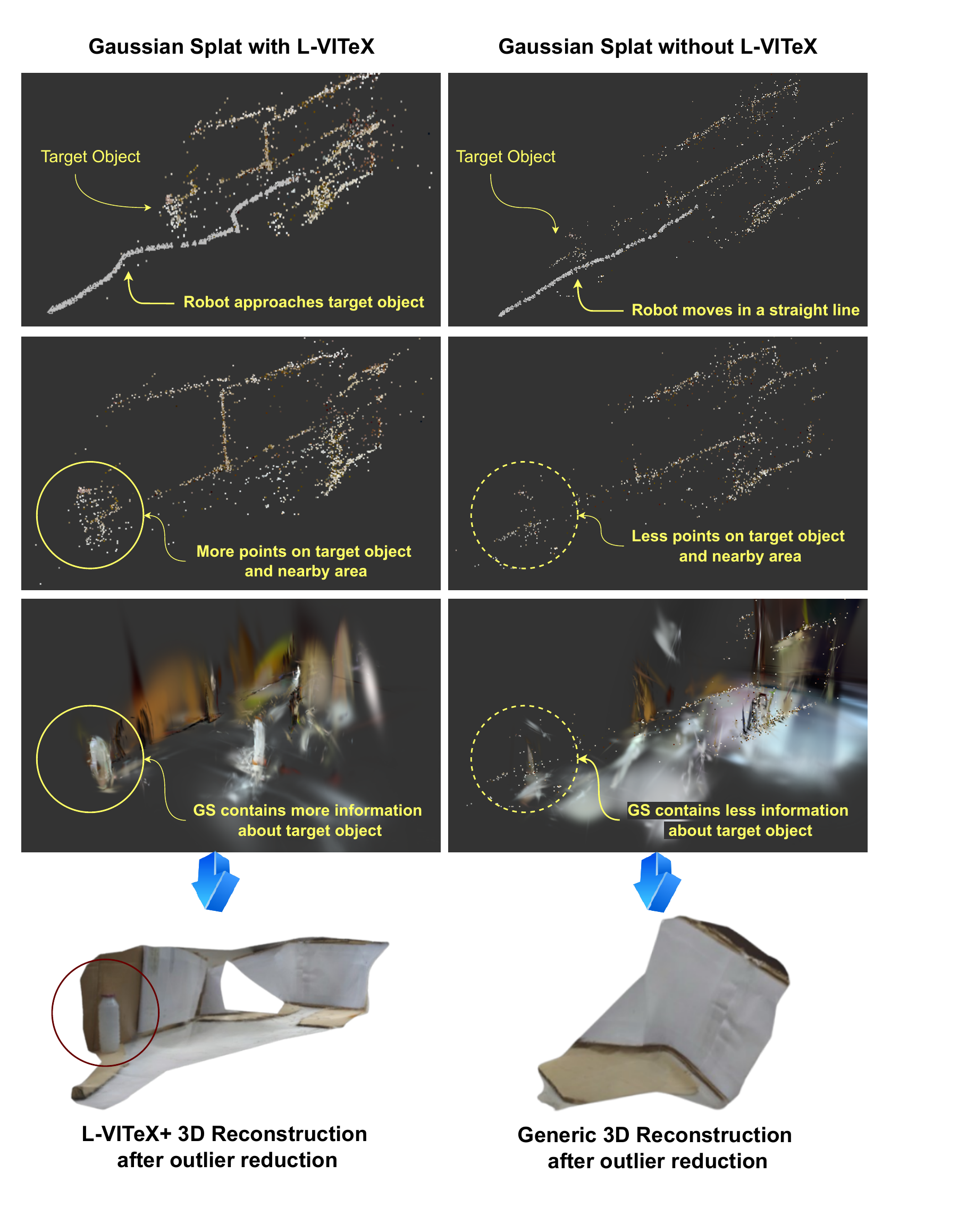}
    \caption{More emphasis resulting in more points for the target object and RoI with L-VITeX resulting in more useful 3D reconstruction through Gaussian Splatting. }
    \label{fig:gs_results}
\end{figure}

Evident from the GS of a portion of the explored tunnel in figure \ref{fig:gs_results}, with our proposed visual heuristic, the robot can detect the target object in the tunnel and approach RoI to get a better look. With this emphasis function, more points are visibly found on the target object and nearby region and the resulting GS-based 3D reconstruction contains more information about the target. On the contract, the lack of attention on the target and RoI is almost absent when the robot simply moves in a straight line and gives almost equal attention to all portions of the tunnel portion.
\subsection{Discussion}
As evident from the results, the FOMO-based object detection works reasonably well in a versatile range of terrains. In addition, the inference latency is also below 1 sec for the input sizes 32x32 and 64x64. In particular, the int8 format with 32x32 input performs with more than 80\% classification accuracy in most cases while having an inference latency as low as 149 ms. the 64x64 input, on the other hand, is the most balanced input size among the three as it achieves higher accuracy than most int8 inferences while also keeping the latency below 1000 seconds, unlike the 96x96 input size. In addition, the input sizes 64x64 and 32x32, especially the latter have very low peak RAM usage (almost half in the case of ESP32-CAM's 520 KB).
Again, from the proof of concept results, we get a visual confirmation that the proposed heuristic can indeed facilitate more useful 3D reconstruction by Gaussian Splat. By closely capturing the target and RoI for more time, we get more frames of the same facilitating SfM to generate more relevant points than wasting resources on background noise.  relevant Therefore, the proposed model can indeed perform near real-time RoI detection for providing visual heuristics in terrain exploration by low-resource explorers.

A closer inspection of the model performance across the datasets in figure \ref{fig:datasets} shows that the model performs best when there is a good separation of background colour against the object colour. For example, the best performance metrics were found for underwater shipwreck detection where the bluish background is clearly separable from the grayish ship debris. In such cases, even the 32x32 input resolution attains very high-performance metrics. On the contrary, the worst performance was found in the case of Martian rock detection. In this case, the oxidized rocks have almost the same colour composition as their background. As a result, with resolution reduction, the rocks, especially the small ones become almost fused with the background and thus remain undetected. Therefore, we can hypothesize from the evidence that, the best-case scenario for the proposed approach is to explore terrains where the RoIs are more contrasting in colour and appearance than their background. 
Another concern about the FOMO-based detection is its merging multiple object centroids into one when they are in the same grid. Thus, the proposed model is not suitable for detected RoI or object counting. However, as the goal of the research is to provide a visual "hunch" of the presence of RoI, counting the same is not generally a priority.

\section{Conclusion}
This research was an exhaustive effort to investigate how far we could push object detection on edge devices for vision-guided terrain exploration. It delved into a comprehensive literature review to sort out the state of the art in visual robot guidance and aimed to provide a lightweight yet high-performance vision heuristic system based on Edge Impulse's FOMO architecture for resource-constrained small explorers and swarms. It demonstrated that for well-separable RoI-background scenarios, the proposed method could be a potential solution.  To consolidate its claim, it also implemented the system into a physical mobile robot to evidently show how it improved vision-based exploration tasks such as 3D reconstruction through Gaussian Splatting. In the future, the authors intend to build upon the proposed L-VITeX system to further improve performance for hard-to-separate RoI-background scenarios while also minimizing current latency margins.

\section*{Acknowledgments}
This research was self-supported completely by the authors. 

\bibliographystyle{ieeetr}  
\bibliography{templateArxiv}

\end{document}